\documentclass{article}

\usepackage{PRIMEarxiv}

\usepackage[utf8]{inputenc} 
\usepackage[T1]{fontenc}    
\usepackage{hyperref}       
\usepackage{url}            
\usepackage{booktabs}       
\usepackage{amsfonts}       
\usepackage{nicefrac}       
\usepackage{microtype}      
\usepackage{lipsum}
\usepackage{fancyhdr}       
\usepackage{graphicx}       
\graphicspath{{media/}}     

\title{Learning Cluster Patterns for Abstractive Summarization
}

\author{
  Sung-Guk Jo\\
  Department of Software Science and Engineering \\
  Kunsan National University \\
   \And
  Jeong-Jae Kim\\
  Graduate School in Cognitive Science \\
  Yonsei University \\
   \And
  Byung-Won On~\thanks{Corresponding author: Byung-Won On (bwon@kunsan.ac.kr)} \\
  Department of Software Science and Engineering \\
  Kunsan National University \\
}

\begin{document}
\maketitle

\begin{abstract}
Nowadays, pre-trained sequence-to-sequence models such as BERTSUM and BART have shown state-of-the-art results in abstractive summarization. In these models, during fine-tuning, the encoder transforms sentences to context vectors in the latent space and the decoder learns the summary generation task based on the context vectors. In our approach, we consider two clusters of salient and non-salient context vectors, using which the decoder can attend more to salient context vectors for summary generation. For this, we propose a novel clustering transformer layer between the encoder and the decoder, which first generates two clusters of salient and non-salient vectors, and then normalizes and shrinks the clusters to make them apart in the latent space. Our experimental result shows that the proposed model outperforms the existing BART model by learning these distinct cluster patterns, improving up to 4\% in ROUGE and 0.3\% in BERTScore on average in CNN/DailyMail and XSUM data sets.
\end{abstract}


\section{Introduction}
\label{sub:intro}

As one of the challenging problems of natural language processing, various methods for automatically summarizing text documents have been actively studied so far. Text summarization are mostly divided into extractive and abstractive approaches. In the extractive approach, a summary is generated by extracting and combining a few important sentences from the original text. On the other hand, the abstractive approach reproduces a short text containing important information, rather than taking part of the original text as it is. 

Since abstractive summarization is a relatively more difficult task than extractive summarization because it must understand the meaning of the original text as well as the natural generation of sentences, most studies have focused on statistical learning-based extractive summarization solutions. However, recently, starting with the abstractive summarization model using the recurrent neural network-based sequence-to-sequence architecture~\cite{rush2015emnlp} and~\cite{nallapati2016conll}, transformer-based pre-trained models such as BERT~\cite{devlin2019naacl} and GPT~\cite{radford2018improving} have been widely used around text summarization areas.

A transformer encoder such as BERT is suitable for extractive summarization that selects a few salient sentences from the original text, but not for abstractive summarization generating a concise text. On the other hand, a transformer decoder such as GPT is suitable for summary generation, but cannot learn bidirectional interactions. To clear such hurdles, state-art-of-the studies on abstractive summarization have mainly focused on pre-trained sequence-to-sequence models such as BERTSUM and BART that combine the transformer encoder and decoder.

In the pre-training step, an original text $d$ is input to both encoder and decoder, which are trained with the corrupted text of $d$. By learning large amounts of text in a corpus, the encoder-decoder model can develop its ability to distinguish between the original and noised text data. In the fine-tuning step, $d$ is input to the encoder and the reference summary to the decoder. The encoder transforms sentences in $d$ to context vectors in the latent semantic space and the decoder learns the summary generation task through the auto-regressive language model objective, based on the context vectors. 

Because a well-written abstractive summary is a short text of the original text without noise information that accounts for most of the original text, handling such a noise information is critical in abstractive summarization. In this work, we view the noise information to non-salient sentences in the original text. For effective summary generation, if the decoder takes a look at cluster information e.g., two clusters of salient and non-salient context vectors during fine-tuning, it can attend more to salient context vectors rather than non-salient ones. 

All sentences composing a text document are relevant with each other regardless of their importance so the corresponding context vectors are closely located to each other in the latent semantic space. In our approach, we force the salient context vectors away from the non-salient ones. As a result, two clusters of salient and non-salient context vectors exist in the latent space. Such clusters are far apart from each other so noise information such as non-salient context vectors is less affected for summary generation.

To use this cluster information, we propose a novel method of learning cluster patterns of context vectors, based on BART, the best one among pre-trained sequence-to-sequence models. We also propose (1) cluster normalization using which the decoder learns cluster patterns more robustly and (2) cluster shrinking that reduces the variance between vectors in a cluster and maximizes the margin between two clusters. In particular, note that our method only transforms existing context vector representation $\vec{v}_i$ to new vector representation $\vec{v}_i^{'}$, which contains clustering information. \textbf{To the best of our knowledge, our work is the first study to use cluster patterns of context vectors for summary generation.} In XSUM and CNN/DailyMail data sets, the experimental result shows that the proposed model improves up to 4\% in ROUGE-1/2/L and 0.3\% in BERTScore on average, compared to the existing BART model.
\section{Main Proposal}
\label{sub:proposal}

\subsection{Background: BART}
\label{subsec:bart}

Abstractive Summarization is based on a language model such as GPT, which learns in the process of predicting the next word with a given word sequence. In this case, words are predicted conditioned on only leftwise context so that it cannot learn in both leftwise and rightwise directions. 

To address this problem,~\cite{devlin2019naacl} proposed the masked language model called BERT. Instead of predicting the word after a given sequence, it learns in the process of first informing the model of the entire sequence and then predicting which word corresponds to the (masked) blank. However, there is a disadvantage that it cannot be easily used for summary generation. 

To improve abstractive summarization,~\cite{lewis2020acl} recently proposed BART, a pre-trained sequence-to-sequence model that combines both BERT as the encoder and GPT as the decoder. BERT and GPT are transformers for bidirectional maksed and auto-regressive language models, respectively. In particular, BART is a denoising autoencoder model. Specifically, it is trained by first corrupting text with one of arbitrary noising functions such as token masking, token deletion, text infilling, sentence permutation, and document rotation, and then learning the model to reconstruct the original text. 

\begin{figure}[tb]
\centering
\begin{tabular}{c}
\mbox{\includegraphics[scale=0.55]{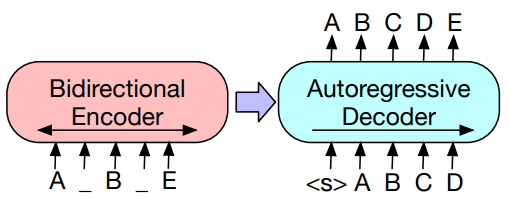}}
\end{tabular}
\caption{Overview of BART~\cite{lewis2020acl}.}
\label{fig:bart}
\end{figure}

As shown in Figure~\ref{fig:bart}, the corrupted text created by one of the noising functions is first encoded with a bidirectional encoder and then a summary is generated with an auto-regressive decoder. For pre-training, the input of the encoder is the corruped text, while the input of the decoder is the original text. For fine-tuning, the input of the encoder is the original text, while the input of the decoder is the reference summary.

In BART, the following noising functions are mainly used.
\begin{itemize}
\item Token Masking (TM): Some tokens are sampled at random and replaced with [MASK] tokens.
\item Token Deletion (TD): Randomly sampled tokens are deleted from the original text.
\item Token Infilling (TI): BART does span masking, not token masking in BERT. For example, a text span of three tokens is replaced with a single [MASK] token.
\item Sentence Permutation (SP): A original text is decomposed into multiple sentences, and then the sentences are shuffled in random order.
\item Document Rotation (DR): Each token in the original text is replaced with the start token of the text. This trains the model to identify the beginning of the text.
\end{itemize}

According to Lewis et al.'s expriemental result, both TI and SP showed the best performance for text summarization. Thus, we used them as the noising function in our experiments.

In addition, Lewis et al. reported that BART outperformed main state-of-the-art summarization models including Lead-3, PTGEN~\cite{see2017acl}, PTGEN+COV~\cite{see2017acl}, UniLM~\cite{dong2019unified}, BERTSUMABS~\cite{liu2019text}, and BERTSUMEXTABS~\cite{liu2019text}. Therefore, we compared our proposed model to BART in this paper.

\subsection{Proposed Method}

Our proposed model for abstractive summarization is based on BART as described in~\ref{subsec:bart}. We extend the existing BART model by adding~\textit{discriminator} $\Delta$ and \textit{clustering transformer} $\tau$. The pre-training step is the same as the existing BART model. However, the fine-tuning step is quite different. Given an original text as input, the discriminator first splits the original text into a set of sentences and then classifies whether each sentence is salient or not. The clustering transformer takes as input context vectors from the encoder in BART and forms two clusters: One is the group of the context vectors corresponding to salient sentences, while the other is the group of the context vectors corresponding to non-salient sentences. The decoder in BART learns these cluster patterns for summary generation and predicts words auto-regressively.

\begin{figure}[tb]
\centering
\begin{tabular}{c}
\mbox{\includegraphics[scale=0.55]{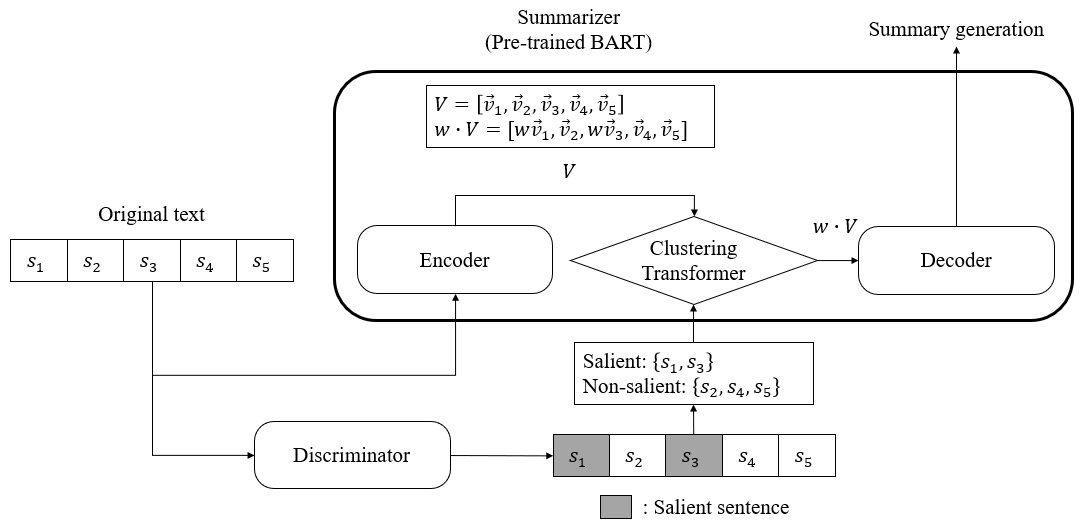}}
\end{tabular}
\caption{Overview of the proposed model.}
\label{fig:proposal}
\end{figure}

Figure~\ref{fig:proposal} illustrates the overview of the proposed model for abstractive summarization. The original text document $d$ consists of five sentences $d = \{s_1, s_2, s_3, s_4, s_5\}$. $\Delta$ takes $d$ as input and classfies whether each sentence is salient or not. For example, if $\Delta
$ classifies $s_1$ and $s_3$ to be salient, while classifying $s_2$, $s_4$, and $s_5$ to be non-salient, we can see two clusters: \{$s_1$, $s_3$\} and \{$s_2$, $s_4$, $s_5$\}. The output of the encoder in BART is a set of the context vectors corresponding to the sentences in $d$. When $\vec{v}_i$ is a context vector corresponding to sentence $s_i$, we can see two clusters of the context vectors: \{$\vec{v}_1$, $\vec{v}_3$\} and \{$\vec{v}_2$, $\vec{v}_4$, $\vec{v}_5$\}.

Based on this clustering information, the clustering transformer converts existing context vectors $\vec{v}_1$, $\vec{v}_2$, $\vec{v}_3$, $\vec{v}_4$, and $\vec{v}_5$ to new context vectors $\vec{v}_1^{'}$, $\vec{v}_2^{'}$, $\vec{v}_3^{'}$, $\vec{v}_4^{'}$, and $\vec{v}_5^{'}$, satisfying the conditions below:
\begin{itemize}
\item $\vec{v}_1^{'} \simeq \vec{v}_3^{'}$,
\item $\vec{v}_2^{'} \simeq \vec{v}_4^{'} \simeq \vec{v}_5^{'}$, and
\item \{$\vec{v}_1^{'}$, $\vec{v}_3^{'}$\} is far apart from \{$\vec{v}_2^{'}$, $\vec{v}_4^{'}$, $\vec{v}_5^{'}$\}.
\end{itemize}

\subsection{Discriminator $\Delta$}

\begin{figure}[tb]
\centering
\begin{tabular}{c}
\mbox{\includegraphics[scale=0.8]{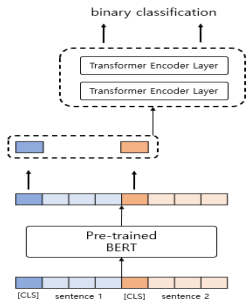}}
\end{tabular}
\caption{Discriminator $\Delta$.}
\label{fig:bertsum}
\end{figure}

\begin{figure*}[tb]
\begin{center}
\begin{tabular}{cc}
\includegraphics[scale=0.3]{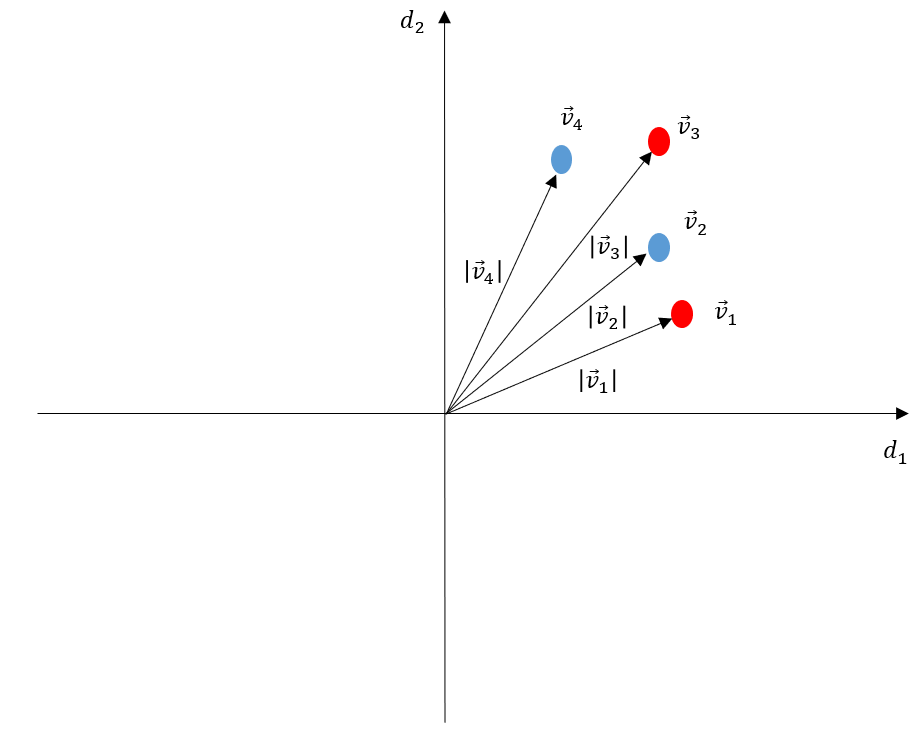} & \includegraphics[scale=0.3]{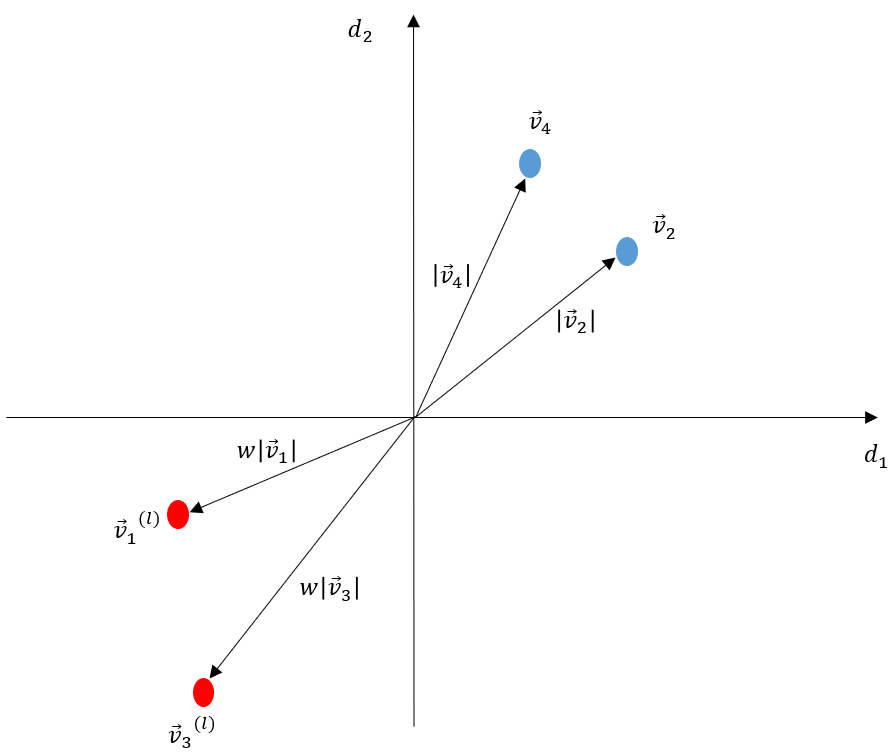} \\
  (a) Context vectors in latent space & (b) Cluster generation\\
\includegraphics[scale=0.3]{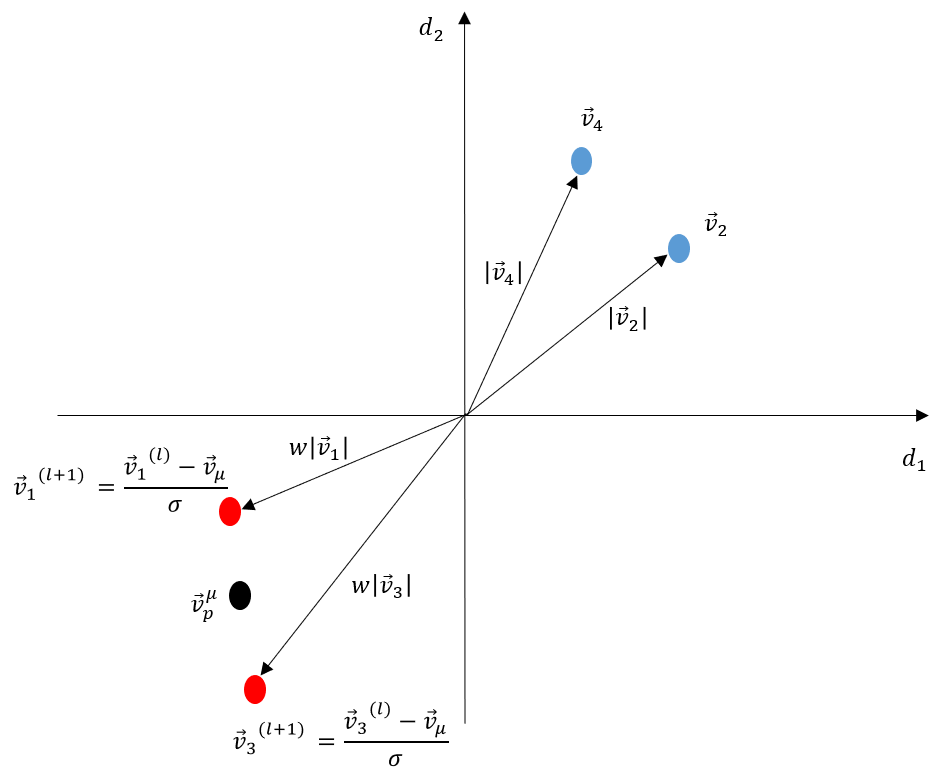} & \includegraphics[scale=0.3]{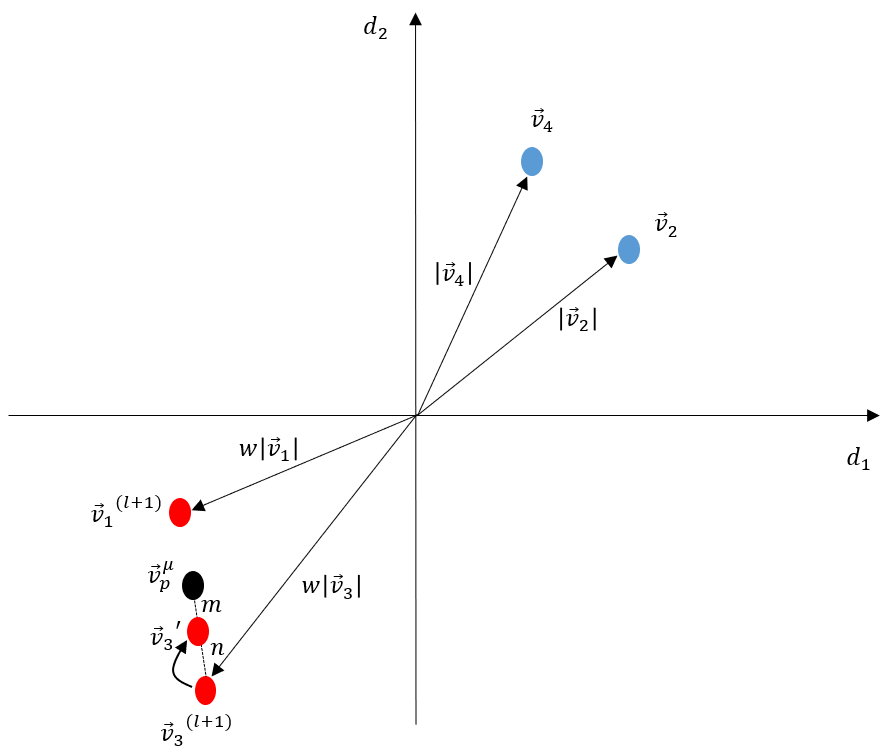} \\
  (c) Cluster normalization & (d) Cluster shrinking \\
\end{tabular}
\caption{Clustering trnasformer $\tau$.}
\label{fig:clustering}
\end{center}
\end{figure*}

For $\Delta$, we use the pre-trained BERT with transformer layers, borrowing similar idea from~\cite{liu2019text}. Figure~\ref{fig:bertsum} shows the neural architecture of $\Delta$ by adding transformer encoders for classification as two layers on top of the existing BERT. The $\Delta$ divides the input text into sentences and adds a [CLS] token at the beginning of each sentence. Among the token vectors through BERT, only [CLS] token vectors are selected and transfered to the transformer encoders. The [CLS] token vectors, which contain the representation of each sentence, are finally output as values of 0 to 1. The higher the output value, the more important the sentence. The sentences with high value are considered to be salient.

\begin{equation}
C_p, C_n \leftarrow  \Delta(d, \lambda)
\end{equation}

In Eq. 1, $C_p$ is the set of salient sentences, while $C_n$ is the set of non-salient sentences. In addition, $\lambda$ is a hyper-parameter controlling the number of salient sentences in $d$. This $\lambda$ value can be different across various data sets. In our experiments, we find the optimal $\lambda$ by investigating ROUGE results in different $\lambda$ values.

\subsection{Clustering Transformer $\tau$}
\label{sec:ct}

The goal of this layer is to transform a context vector $\vec{v}_i$ from the encoder neural model in BART to a new vector representation $\vec{v}_i^{'}$ including clustering information. Please take a look at the following Eq. 2.

\begin{equation}
\vec{v}_i^{'} \leftarrow \tau(\vec{v}_i)
\end{equation}
, where $\tau()$ is the function for the clustering transformer.

Figure~\ref{fig:clustering}(a) depicts four contextualized vectors $\vec{v}_1$, $\vec{v}_2$, $\vec{v}_3$, and $\vec{v}_4$ in the latent space~\footnote{We simplify the number of vector dimensions as two for intuitively understanding. In the coordinates, $d_1$ and $d_2$ are the first dimension and the second one, respectively.}. Supposing that a given original text $d$ consists of four sentences $s_1$, $s_2$, $s_3$, and $s_4$, e.g., $d=\{s_1, s_2, s_3, s_4\}$, $\vec{v}_{i}$ is the context vector of $s_{i}$ and |$\vec{v}_{i}$| is the length of  $\vec{v}_{i}$. In the figure, the red vectors (e.g., $\vec{v}_{1}$ and $\vec{v}_{3}$) are lableled as salient sentences from $\Delta$, while the blue ones (e.g., $\vec{v}_{2}$ and $\vec{v}_{4}$) as non-salient ones.

\textbf{Cluster Generation}: To cluster salient and non-salient sentences, the representation of the context vectors corresponding to the salient sentences is recalculated by the following equation.

\begin{displaymath}
\vec{v}_i^{(l)} = \left\{ \begin{array}{ll}
  \tau_1{(w \cdot \vec{v}_{i})} & \textrm{if $\vec{v}_{i} \in C_p$}\\
  \tau_1{(\vec{v}_{i})} & \textrm{if $\vec{v}_{i} \in C_n$}
 \end{array} \right.
\end{displaymath}

, where $w$ is a certain weight value that is a hyper-parameter and we used the optimal $w$ through a series of experiments with different $w$ values. As shown in Figure~\ref{fig:clustering}(b), when $w = -1$, $\vec{v}_{i}^{(l)}$ is located in the opposite direction of $\vec{v}_{i}$ in the latent space. In the figure, context vectors corresponding to non-salient sentences exist in the first quadrant of latent space coordinates, and ones corresponding to salient sentences move to the third quadrant. As a result, two clusters can be formed in the latent space. Now vectors $\vec{v}_1^{(l)}$ and $\vec{v}_3^{(l)}$ are changed to new vector representation containing clustering information.

\textbf{Cluster Normalization}: In addition, $C_p^{'} = \{ \vec{v}_1^{(l)}, \vec{v}_3^{(l)} \}$ can be further normalized into $C_p^{''} = \{ \vec{v}_1^{(l+1)}, \vec{v}_3^{(l+1)} \}$ using Eq. 3 below.

\begin{equation}
\vec{v}_i^{(l+1)} = \tau_2{(\frac{\vec{v}_i^{(l)} - \vec{v}_{p}^{\mu}}{\sigma})}
\end{equation}

, where $\vec{v}_p^{\mu}$ is the mean vector, shown as the black vector in Figure~\ref{fig:clustering}(c), and $\sigma$ is the standard deviation in $C_p^{'}$. $C_n^{'} = \{ \vec{v}_2^{(l)}, \vec{v}_4^{(l)} \}$ is also normalized into $C_n^{''} = \{ \vec{v}_2^{(l+1)}, \vec{v}_4^{(l+1)} \}$. We expect that thransformation of the vector representation through Eq. 3 will enable the decoder model in BART to learn cluster patterns more robustly.

\textbf{Cluster Shrinking}: Our another approach is to forcibly shrink the size of clusters ($C_p^{'}$ and $C_n^{'}$) that have already been created by Eq. 3. We first obtain the mean vectors $\vec{v}_{p}^{\mu}$ in $C_p^{'}$ and $\vec{v}_{n}^{\mu}$ in $C_n^{'}$, respectively. Next, for each cluster (e.g., $C_p^{'}$), we compute Euclidean distances between $\vec{v}_{p}^{\mu}$ and $\vec{v}_i^{(l+1)} \in C_p^{'}$. For instance, after the distance $dist(\vec{v}_p^{\mu}, \vec{v}_3^{(l+1)})$ is computed, it is indicated by the dotted line in Figure~\ref{fig:clustering}(d). Finally, we divide the distance by the ratio $m$ to $n$. If $m=n=1$, we move $\vec{v}_3^{(l+1)}$ to the center point of the distance.

\begin{equation}
\vec{v}_i^{'} = \tau_3{(\frac{dist(\vec{v}_p^{\mu}, \vec{v}_i^{(l+1)})}{2})}~\textnormal{ if }m=n=1
\end{equation}

, where $m$ and $n$ are hyper-parameters and we discovered the optimal ratio value through a series of experiments with different ratios of $m$ to $n$. This approach can make it easier for the decoder model in BART to learn distinct cluster patterns because it reduces the variance between vectors in a cluster by making the vectors in the cluster close together at a constant rate.
\section{Experimental Set-up}
\label{subsec:setup}

For the experiment, we first implemented the discriminator and the clustering transformer using Python and then combined them into existing BART base model. To execute the BART model, we set 0.1 to dropout rate, 1 to batch size, 3e-5 to learning rate, 2 to gradient accumulation, and 0.1 to gradient clipping. As the hyper-parameters for summary generation, we set 6 to the number of beams, 62 to max length, 3 to max repeat $n$-gram, and TRUE to early stopping. We used 1,024 tokens as the maximum input size of BART.

In addition to the hyper-parameters of the existing BART model, the hyper-parameters of the proposed model are $\lambda$, $w$, $m$, and $n$, where $\lambda$ is the number of salient sentences; $w$ is the weight value for forming the cluster of salient sentences; and the ratio of $m$ to $n$ for cluster shrinking. We selected best hyper-parameters through a series of experiments with different values.

\begin{table*}
\centering
\begin{tabular}{lcc}
\hline
\textbf{} & \textbf{BART} & \textbf{Proposed model}\\
\hline
ROUGE-1 & 36.48 & 37.81 \\
ROUGE-2 & 14.36 & 15.07 \\
ROUGE-L & 29.18 & 30.27 \\
ROUGE-avg & 26.67 & 27.72 \\
BERTScore & 90.06 & 90.39 \\\hline
\end{tabular}
\begin{tabular}{lcc}
\hline
\textbf{} & \textbf{BART} & \textbf{Proposed model}\\
\hline
ROUGE-1 & 41.11 & 41.99 \\
ROUGE-2 & 19.54 & 20.01 \\
ROUGE-L & 38.18 & 38.85 \\
ROUGE-avg & 32.94 & 33.62 \\
BERTScore & 88.21 & 88.30 \\\hline
\end{tabular}
\caption{ROUGE and BERTScore values of BART and the proposed model (Left: XSUM and Right: CNN/DailyMail).}
\label{tab:rougeBertscore}
\end{table*}

To evaluate the proposed model for abstractive summarization, we used two distinct data sets: One is CNN/DailyMail~\cite{nallapati2016conll} and the other is XSUM~\cite{narayan2018emnlp}, which are main benchmark data sets in the text summarization field. It is known that the characteristics of the two data sets are slightly different. Reference summaries look like extractive style summaries on the CNN/DailyMail data set. On the other hand, in the XSUM data set, the reference summaries are extremely abstractive summaries, consisting of one or two sentences.

We used ROUGE 1/2/L, which have been widely employed as the evaluation metric for text summarization~\cite{lin2004acl}. Such metrics tend to estimate lexical similarity between two sequences by counting matched tokens between them. However, in case of tokens with same meaning but yet different spellings, such lexical-based similarity measures do not work effectively. To surmount this limitation, we also used BERTScore~\cite{zhang2020iclr} to compute the semantic similarity between two sequences.

All models were in standalone executed in a high-performance workstation server with Intel Xeon Scalable Silver 4414 2.20GHz CPU with 40 cores, 24GB RAM, 1TB SSD, and GEFORCE RTX 2080 Ti D6 11GB BLOWER with 4,352 CUDA cores, 12GB RAM, and 7 GBPS memory clock.

\section{Experimental Results}

\subsection{Comparison of Proposed Model to BART}

We first tested BART and the proposed model with XSUM and CNN/DailyMail data sets, respectively, and then measured ROUGE-1/2/L and BERTScore values. Table~\ref{tab:rougeBertscore} summarizes the results in order to compare the preformance of the proposed model with the existing BART model. After we ran the proposed model with different hyper-parameters such as $\lambda$, $w$, and the ratio of $m$ to $n$ and chose them showing the best ROUGE and BERTScore, we obtained $\lambda = 0.5$, $w = -1.0$, $m = 3$, and $n = 1$ in XSUM data set and $\lambda = 0.6$, $w = -1.0$, $m = 6$, and $n = 1$ in CNN/DailyMail data set. 

Compared with BART, the proposed model largely improves ROUGE-1/2/L by 4\% and BERTScore by 0.3\% on average in XSUM and improves ROUGE-1/2/L by 2\% and BERTScore by 0.001\% on average in CNN/DailyMail. The reason why the proposed model shows higher performance than the existing BART model is that the decoder generates a summary after learning cluster patterns made by the proposed clustering transformer, which creates two clusters for salient and non-salient sentences and increases the margin between the two clusters. For fine-tuning, the decoder is less affected by noise information such as non-salient sentences, attending to the clusters including salient sentences. As a result, it can generate a summary including the main information of the original text.

Interestingly, regardless of BART and the proposed model, all the ROUGE scores in the CNN/DailyMail data set are slightly higher than those in the XSUM data set. For example, the ROUGE-avg score of the proposed model is 27.72 in XSUM, whereas it is 33.62 in CNN/DailyMail. In contrast, all the BERTScore values in the CNN/DailyMail data set are slightly lower than those in the XSUM data set. For instance, the BERTScore value of the proposed model is 90.39 in XSUM, whereas it is 88.30 in CNN/DailyMail. This difference is due to the characteristics of the two data sets. Actually, most of the reference summaries of the XSUM data set were paraphrased from the original text as an extreme abstractive summary consisting of one or two sentences. On the other hand, each reference summary of the CNN/DailyMail data set is the set of salient sentences taken directly from the original text.

\subsection{Results of Different Weight Values}

\begin{table}
\centering
\begin{tabular}{lcc}
\hline
\textbf{$w$} & \textbf{ROUGE-1/2/L} & \textbf{BERTScore}\\
\hline
-2.0 & 37.34/14.90/29.90 & 90.24 \\
-1.5 & 37.46/14.87/29.96 & 90.26 \\
-1.0 & 37.54/14.96/30.04 & 92.29 \\
 1.5 & 37.06/14.71/29.40 & 90.16 \\
 2.0 & 37.04/14.61/29.28 & 90.10 \\\hline
\end{tabular}
\caption{ROUGE-avg and BERTScore values of the proposed model according to different weight values.}
\label{tab:weight}
\end{table}

As shown in Section~\ref{sec:ct}, by scaling a context vector by a certain weight value $w$, context vectors corresponding to salient and non-salient sentences are clustered in the latent space. If $w$ is between 0 and 1, the length of new vector $\vec{v}_i^{(l)}$ is reduced, compared to the given context vector $\vec{v}_i$, but the direction is the same. Furthermore, if $w$ is greater than 1, the length of $\vec{v}_i^{(l)}$ increases but the direction is also the same. On the other hand, if $w$ is between 0 and -1, the length of $\vec{v}_i^{(l)}$ is small, indicating that it is in the opposite direction, compared to $\vec{v}_i$. If $w$ = -1, new vector $\vec{v}_i^{(l)}$ has the same length but the opposite direction.

If the cluster of salient sentence vectors is far from the cluster of non-salient ones in the latent space, the decoder in BART can focus mainly on learning the salient sentence vectors out of non-salient ones. For this reason, selecting an appropriate weight value $w$ is important in the proposed model. In this experiment, we performed the proposed model with different weight values $w$ = \{-2.0, -1.5, -1.0, 1.5, 2.0\} in the XSUM data set and measured its ROUGE-avg and BERTScore values as summarized in Table~\ref{tab:weight}.

When $w$ = -1.0, the proposed model shows the best results in both ROUGE-avg and BERTScore. These results are reasonable because of the relative position between the two clusters in the latent space. For any vector $\vec{v}_i$, the proposed model can generate the best summary when new vector $\vec{v}_i^{(l)}$ has the same length as $\vec{v}_i$ and is located in the latent space in the opposite direction. In other words, when $w$ = -1.0, two clusters of salient and non-salient sentence vectors are well distinguished, but when $w$ = 0.8, two clusters are close to each other, so they may not be well distinguished in the latent space.

\begin{table}
\centering
\begin{tabular}{lcc}
\hline
\textbf{$\lambda$} & \textbf{ROUGE-1/2/L} & \textbf{BERTScore}\\
\hline
0.2 & 37.84/15.26/30.20 & 90.33 \\
0.3 & 37.68/14.99/30.07 & 90.36 \\
0.4 & 37.74/14.98/30.04 & 90.30 \\
0.5 & 37.81/15.07/30.27 & 90.39 \\
0.6 & 37.55/15.10/30.05 & 90.27 \\
0.7 & 37.40/14.91/29.96 & 90.30 \\\hline
\end{tabular}
\caption{ROUGE-avg and BERTScore values of the proposed model according to different $\lambda$ values.}
\label{tab:lambda}
\end{table}

\subsection{Results of Number of Salient Sentences}

As an input to the discriminator $\Delta$, $\lambda$, one of the hyper-parameters in the proposed model, controls the number of salient sentences. Supposing an original text $d = \{s_1, s_2, ..., s_{10}\}$, $\Delta$ extracts three sentences as salient sentences in $d$ if we set $\lambda$ to 0.3. In this work, we evaluated how well the proposed model generates a summary according to the number of salient sentences per $d$. For this, we experimented the proposed model with different $\lambda$ values in the XSUM data set and compared the results of ROUGE-1/2/L and BERTScore.

Table~\ref{tab:lambda} shows the results according to different $\lambda$ values. In this experiment, we used $w$ = -1.0 and the ratio of $m = 3$ to $n = 1$ in cluster shrinking. The experimental results are similar to each other. However, when $\lambda = 0.2$, the ROUGE-1/2/L score is marginally better, and when $\lambda = 0.5$, the BERTScore value is slightly higher than the others. Overall, the ROUGE and BERTScore values seem to increase slightly until $\lambda = 0.5$, and then, from $\lambda = 0.5$, those values tend to decrease slightly. From these experimental results, we can figure out that $\lambda = 0.5$ is better for summary generation. Similar results were also obtained for the CNN/DailyMail data set, which is omitted here due to space limitations.

\subsection{Results of Ratio of $m$:$n$ for Cluster Shrinking}

\begin{table}
\centering
\begin{tabular}{lcc}
\hline
\textbf{$m$:$n$} & \textbf{ROUGE-1/2/L} & \textbf{BERTScore}\\
\hline
1:1 & 37.49/14.67/30.00 & 90.30 \\
2:1 & 37.64/15.03/30.08 & 90.34 \\
3:1 & 37.81/15.07/30.27 & 90.39 \\
4:1 & 37.47/14.83/29.94 & 90.30 \\
5:1 & 37.48/14.99/30.11 & 90.31 \\
6:1 & 37.60/15.09/30.18 & 90.36 \\\hline
\end{tabular}
\caption{ROUGE and BERTScore values of the proposed model according to different ratios of $m$ to $n$ values.}
\label{tab:mn}
\end{table}

To reduce the size of clusters, we first compute the mean vector in each cluster and then find the distance between the mean vector and any vector belonging to the cluster. Subsequently, we divide this distance by the ratio of $m$ to $n$ and move the existing vector to the center position between $m$ and $n$. The purpose of this approach is to distinguish salient sentence vectors from non-salient ones well. The size of clusters formed in the latent space can be reduced by maximizing the margin between the two clusters of salient and non-salient vectors. Through this clustering shrinking method, the decoder in BART are likely to learn cluster patterns easily.

Table~\ref{tab:mn} shows the results of the proposed model according to different ratios of $m$ to $n$ values, where $w$ and $\lambda$ are set to -1.0 and 0.5. In the XSUM data set, both the ROUGE-1/2/L and BERTScore values are the highest when $m = 3$ and $n = 1$. On the other hand, in the case of the CNN/DailyMail data set, best ROUGE and BERTScore results are shown when the ratio of $m = 6$ to $n = 1$. For instance, the ROUGE-1/2/L and BERTScore values of the proposed model are 41.99, 20.01, 38.85, and 88.30, respectively. This indicates that the ratio of $m$ to $n$ depends on the characteristics of the given summarization data set for cluster shrinking.

\subsection{Discussion of Generated Summaries}

\begin{table}[h]
\centering
\begin{tabular}{l}
\hline
\textbf{Original text} \\ \hline
A spokesman for Palm Beach Gardens police in Florida confirmed to the BBC they were investigating a fatal crash \\
involving the Grand Slam champion. A man was taken to hospital after the accident on 9 June and died two weeks \\
later from his injuries, he said. According to TMZ, which broke the story, police believe the seven--time Grand \\
Slam champion was at fault. But a lawyer from Williams said it was an ``unfortunate accidient''. The man who \\
died, Jerome Barson, was travelling with his wife who was driving their vehicle through an intersection when the \\
accident happened. Williams's car suddenly darted into their path and was unable to clear the junction in time due \\
to traffic jams, according to witness statements in a police report obtained by US media. Mrs Barson was also taken \\
to hospital but survived. ``Williams is at fault for violating the right of way of the other driver.'' the report said, \\
adding that there were no other factors like drugs, alcohol or mobile phone distractions. The 37--year--old tennis \\
star reportedly told police she did not see the couple's car and she was driving slowly. Police spokesman Major Paul \\
Rogers said police were investigating whether the incident was connected to Mr Barson's death. William's lawyer \\
Malcolm Cunningham told CNN in a statement: ``Ms Williams entered the intersection on a green light. The police \\
report estimates that Ms Williams was travelling at 5mph when Mrs Barson crashed into her. ``Authorities did not \\
issue Ms Williams with any citations or traffic violations. This is an unfortunate accident and Venus expresses her \\
deepest condolences to the family who lost a loved one.'' Next week, Williams is due to play at Wimbledon in \\
London, where she is seeded 10th. \\ \hline

\textbf{Reference summary} \\ \hline
US tennis star Venus Williams has been involved in a car accident that led to the death of a 78--year--old man. \\ \hline
\textbf{Summary generated by BART}\\ \hline
US tennis star \underline{Serena Williams} is at fault in the death of a man who died in a crash, police say.\\ \hline
\textbf{Summary generated by the proposed model}\\ \hline
\underline{Venus Williams} has been arrested on suspicion of causing the death of a man who was killed in a car crash.\\  \hline
\end{tabular}
\caption{An example of summaries generated by BART and the proposed model.}
\label{tab:example}
\end{table}

Table~\ref{tab:example} shows an example of two summaries generated by both BART and the proposed model. The actual original text has a long text span, but due to space limitations, we omitted a significant amount of text in the table. As the reference summary, the main point summarized by a human evaluator is that US tennis star Venus Williams was involved in a car accident with the death of a man. Both summaries generated by BART and the proposed model focus on a car accident with a famous tennis player in US. The BART and proposed models all summarized the original text well as a human-written abstract usually does. In addition, the generated summaries are natural, fluent, and no errors in grammar. 

However, the BART model summarized inccorrectly, while the proposed model did correctly. The US tennis star is not Serena Williams but Venus Williams in the car accident. As we can be seen from these results, learning distinct cluster patterns in the latent space prevents the sequence-to-sequence model for abstractive summarization from inaccurately reproducing factual details.

\section{Related Work}

For abstractive summarization,~\cite{rush2015emnlp} and~\cite{nallapati2016conll} first proposed the encoder-decoder attention model.~\cite{see2017acl} improved the neural model of~\cite{nallapati2016conll} by copying words from an input text through the pointer network, which can handle out-of-vocabulary words, while generating words from the generator network. They also proposed a coverage mechanism that tracks and controls coverage of the input text in order to eliminate repetition. With the great success of the transformer model in natural language generation,~\cite{dong2019unified} and~\cite{liu2019text} proposed the pre-trained model for abstractive summarization. In particular,~\cite{dong2019unified} proposed an unified pre-trained language model (UniLM) that is a multi-layer transformer model as backbone network, jointly pre-trained by various unsupervised language modeling objectivites such as (1) bidirectional LM, (2) right-to-left or left-to-right undirectional LM, and sequence-to-sequence LM, sharing the same parameters.~\cite{liu2019text} used a encoder-decoder model, in which a document-level encoder using BERT is pre-trained on large amounts of text. They also proposed a new fine-tuning schedule to alleviate the mismatch between the encoder and the decoder with BERTSUMABS and BERTSUMEXTABS, which are the baseline and the two-stage fine-tuned models. Recently,~\cite{lewis2020acl} proposed the BART model that consists of BERT as the encoder and GPT as the decoder. The BART model is similar to the BERTSUM proposed in~\cite{liu2019text}, except that the encoder can perform the masking task through various denoising functions.

Besides these pre-trained models, various models using reinforcement learning~\cite{chen2018acl}~\cite{paulus2018iclr}, topic model~\cite{wang2020emnlp}~\cite{cui2021acl}, multimodal information~\cite{yu2021emnlp}, attention head masking~\cite{cao2021naacl}, information theory~\cite{xiao2020emnlp}, extraction-and-paraphrasing~\cite{nikolov2020lrec}, entity aggregation~\cite{jumel2020emnlp}, factuality consistency~\cite{cao2020emnlp}~\cite{dong2020emnlp}~\cite{durmus2020acl}~\cite{kryscinski2020emnlp}~\cite{xu2020acl}~\cite{maynez2021acl}, deep communicating agents~\cite{celikyilmaz2021naacl}, sentence correspondence~\cite{lebanoff2020acl}, graph~\cite{huang2020aaai}~\cite{jin2020acl}~\cite{li2020acl}, and bottom-up approach~\cite{gehrmann2020emnlp} were proposed in abstractive summarization. Because these models are not directly related to our proposed model, we do not compared with them in this paper.

\section{Conclusion}

To improve abstractive summarization, we propose a new pre-trained sequence-to-sequence model containing a discriminator and a clustering transformer layer between the encoder and the decoder. During fine-tuning, the discriminator extracts salient sentences from a given original text. The clustering transformer generates two clusters of salient and non-salient context vectors from the encoder, and normalizes and shrinks the clusters to better distinguish the two clusters. Our experiment result shows that the proposed method outperforms the existing BART model in two summarization benchmark data sets.

\section{Acknowledge}

We thank the anonymous reviewers for their feedback. We gratefully acknowledge the support of the National Research Foundation of Korea (NRF) Grant by the Korean Government through MSIT under Grant NRF-2019R1F1A1060752.

\bibliographystyle{unsrt}  
\bibliography{references}

\begin{thebibliography}{10}

\bibitem{rush2015emnlp}
Alexander~M. Rush, Sumit Chopra, and Jason Weston.
\newblock A neural attention model for abstractive sentence summarization.
\newblock In {\em Proceedings of the 2015 Conference on Empirical Methods in
  Natural Language Processing}, pages 379--389, 2015.

\bibitem{nallapati2016conll}
Ramesh Nallapati, Bowen Zhou, Cicero~dos Santos, Caglar Gulcehre, and Bing
  Xiang.
\newblock Abstractive text summarization using sequence-to-sequence rnns and
  beyond.
\newblock In {\em Proceedings of the SIGNLL on Computational Natural Language
  Learning}, pages 1--12, 2016.

\bibitem{devlin2019naacl}
Jacob Devlin, Ming-Wei Chang, Kenton Lee, and Kristina Toutanova.
\newblock Bert: Pre-training of deep bidirectional transformers for language
  understanding.
\newblock In {\em Proceedings of the 2019 Conference of the North American
  Chapter of the Association for Computational Linguistics: Human Language
  Technologies}, pages 4171--4186, 2019.

\bibitem{radford2018improving}
Alec Radford, Karthik Narasimhan, Tim Salimans, and Ilya Sutskever.
\newblock Improving language understanding by generative pre-training.
\newblock 2018.

\bibitem{lewis2020acl}
Mike Lewis, Yinhan Liu, Naman Goyal, Marjan Ghazvininejad, Abdelrahman Mohamed,
  Omer Levy, Ves Stoyanov, and Luke Zettlemoyer.
\newblock Bart: Denoising sequence-to-sequence pre-training for natural
  language generation, translation, and comprehension.
\newblock {\em arXiv preprint arXiv:1910.13461}, 2019.

\bibitem{see2017acl}
Abigail See, Peter~J. Liu, and Christopher~D. Manning.
\newblock Get to the point: Summarization with pointer-generator networks.
\newblock In {\em Proceedings of the 55th Annual Meeting of the Association for
  Computational Linguistics}, pages 1073--1083, 2017.

\bibitem{dong2019unified}
Li~Dong, Nan Yang, Wenhui Wang, Furu Wei, Xiaodong Liu, Yu~Wang, Jianfeng Gao,
  Ming Zhou, and Hsiao-Wuen Hon.
\newblock Unified language model pre-training for natural language
  understanding and generation.
\newblock In {\em arXiv preprint arXiv: 1905.03197}, pages 1--14, 2019.

\bibitem{liu2019text}
Yang Liu and Mirella Lapata.
\newblock Text summarization with pretrained encoders.
\newblock In {\em arXiv preprint arXiv: 1908.08345}, pages 1--11, 2019.

\bibitem{narayan2018emnlp}
Shashi Narayan, Shay~B. Cohen, and Mirella Lapata.
\newblock Don't give me the details, just the summary! topic-aware
  convolutional neural networks for extreme summarization.
\newblock In {\em Proceedings of the 2018 Conference on Empirical Methods in
  Natural Language Processing}, pages 1--11, 2018.

\bibitem{lin2004acl}
Chin-Yew Lin.
\newblock Rouge: A package for automatic evaluation of summaries.
\newblock In {\em Proceedings of the 42nd Annual Meeting of the Association for
  Computational Linguistics}, pages 1--8, 2004.

\bibitem{zhang2020iclr}
Tianyi Zhang, Varsha Kishore, Felix Wu, Kilian~Q. Weinberger, and Yoav Artzi.
\newblock Bertscore: Evaluating text generation with bert.
\newblock In {\em Proceedings of the 8th International Conference on Learning
  Representations}, pages 1--43, 2020.

\bibitem{chen2018acl}
Yen-Chun Chen and Mohit Bansal.
\newblock Fast abstractive summarization with reinforce-selected sentence
  rewriting.
\newblock In {\em Proceedings of the 56nd Annual Meeting of the Association for
  Computational Linguistics}, pages 675--686, 2018.

\bibitem{paulus2018iclr}
Romain Paulus, Caiming Xiong, and Richard Socher.
\newblock A deep reinforced model for abstractive summarization.
\newblock In {\em Proceedings of the 2018 International Conference on Learning
  Representations}, pages 1--13, 2018.

\bibitem{wang2020emnlp}
Zhengjue Wang, Zhibin Duan, Hao Zhang, Chaojie Wang, Long Tian, Bo~Chen, and
  Mingyuan Zhou.
\newblock Friendly topic assistant for transformer based abstractive
  summarization.
\newblock In {\em Proceedings of the 2020 Conference on Empirical Methods in
  Natural Language Processing}, pages 485--497, 2020.

\bibitem{cui2021acl}
Peng Cui and Le~Hu.
\newblock Topic-guided abstractive multi-document summarization.
\newblock In {\em Proceedings of the 2021 Annual Meeting of the Association for
  Computational Linguistics}, pages 1463--1472, 2021.

\bibitem{yu2021emnlp}
Tiezheng Yu, Wenliang Dai, Zihan Liu, and Pascale Fung.
\newblock Vision guided generative pre-trained language models for multimodal
  abstractive summarization.
\newblock In {\em Proceedings of the 2021 Conference on Empirical Methods in
  Natural Language Processing}, pages 3995--4007, 2021.

\bibitem{cao2021naacl}
Shuyang Cao and Lu~Wang.
\newblock Attention head masking for inference time content selection in
  abstractive summarization.
\newblock In {\em Proceedings of the 2021 Conference of the North American
  Chapter of the Association for Computational Linguistics}, pages 5008--5016,
  2021.

\bibitem{xiao2020emnlp}
Liqiang Xiao, Lu~Wang, Hao He, and Yaohui Jin.
\newblock Modeling content importance for summarization with pre-trained
  language models.
\newblock In {\em Proceedings of the 2020 Conference on Empirical Methods in
  Natural Language Processing}, pages 3606--3611, 2020.

\bibitem{nikolov2020lrec}
Nikola~I. Nikolov and H.~R. Hahnloser.
\newblock Abstractive document summarization without parallel data.
\newblock In {\em Proceedings of the 12th Conference on Language Resources and
  Evaluation}, pages 6638--6644, 2020.

\bibitem{jumel2020emnlp}
Clement Jumel and Annie Louis.
\newblock Tesa: A task in entity semantic aggregation for abstractive
  summarization.
\newblock In {\em Proceedings of the 2020 Conference on Empirical Methods in
  Natural Language Processing}, pages 8031--8050, 2020.

\bibitem{cao2020emnlp}
Meng Cao, Yue Dong, Jiapeng Wu, and Jackie Chi~Kit Cheung.
\newblock Factual error correction for abstractive summarization models.
\newblock In {\em Proceedings of the 2020 Conference on Empirical Methods in
  Natural Language Processing}, pages 6251--6258, 2020.

\bibitem{dong2020emnlp}
Yue Dong, Shuohang Wang, Zhe Gan, Yu~Cheng, Jackie Chi~Kit Cheung, and Jingjing
  Liu.
\newblock Multi-fact correction in abstractive text summarization.
\newblock In {\em Proceedings of the 2020 Conference on Empirical Methods in
  Natural Language Processing}, pages 9320--9331, 2020.

\bibitem{durmus2020acl}
Esin Durmus, He~He, and Mona Diab.
\newblock Feqa: A question answering evaluation framework for faithfulness
  assessment in abstractive summarization.
\newblock In {\em Proceedings of the 2020 Annual Meeting of the Association for
  Computational Linguistics}, pages 5055--5070, 2020.

\bibitem{kryscinski2020emnlp}
Wojciech Kryscinski, Bryan McCann, Caiming Xiong, and Richard Socher.
\newblock Evaluating the factual consistency of abstractive text summarization.
\newblock In {\em Proceedings of the 2020 Conference on Empirical Methods in
  Natural Language Processing}, pages 9332--9346, 2020.

\bibitem{xu2020acl}
Xinnuo Xu, Ondrej Dusek, Jingyi Li, Verena Rieser, and Ioannis Konstas.
\newblock Fact-based content weighting for evaluating abstractive
  summarization.
\newblock In {\em Proceedings of the 2020 Annual Meeting of the Association for
  Computational Linguistics}, pages 5071--5081, 2020.

\bibitem{maynez2021acl}
Joshua Maynez, Shashi Narayan, Bernd Bohnet, and Ryan McDonald.
\newblock On faithfulness and factuality in abstractive summarization.
\newblock In {\em Proceedings of the 58th Annual Meeting of the Association for
  Computational Linguistics}, pages 1906--1919, 2021.

\bibitem{celikyilmaz2021naacl}
Asli Celikyilmaz, Antoine Bosselut, Xiaodong He, and Yejin Choi.
\newblock Deep communicating agents for abstractive summarization.
\newblock In {\em Proceedings of the 2018 Conference of the North American
  Chapter of the Association for Computational Linguistics}, pages 1662--1675,
  2018.

\bibitem{lebanoff2020acl}
Logan Lebanoff, John Muchovej, Franck Dernoncourt, Doo~Soon Kim, Lidan Wang,
  Walter Chang, and Fei Liu.
\newblock Understanding points of correspondence between sentences for
  abstractive summarization.
\newblock In {\em Proceedings of the 2020 Annual Meeting of the Association for
  Computational Linguistics}, pages 191--198, 2020.

\bibitem{huang2020aaai}
Luyang Huang, Lingfei Wu, and Lu~Wang.
\newblock Knowledge graph-augmented abstractive summarization with
  semantic-driven cloze reward.
\newblock In {\em Proceedings of the AAAI Conference on Artificial
  Intelligence}, pages 7919--7926, 2020.

\bibitem{jin2020acl}
Hanqi Jin, Tianming Wang, and Xiaojun Wan.
\newblock Multi-granularity interaction network for extractive and abstractive
  multi-document summarization.
\newblock In {\em Proceedings of the 2020 Annual Meeting of the Association for
  Computational Linguistics}, pages 6244--6254, 2020.

\bibitem{li2020acl}
Wei Li, Xinyan Xiao, Jiachen Liu, Hua Wu, Haifeng Wang, and Junping Du.
\newblock Leveraging graph to improve abstractive multi-document summarization.
\newblock In {\em Proceedings of the 2020 Annual Meeting of the Association for
  Computational Linguistics}, pages 1--12, 2020.

\bibitem{gehrmann2020emnlp}
Sebastian Gehrmann, Yuntian Deng, and Alexander Rush.
\newblock Bottom-up abstractive summarization.
\newblock In {\em Proceedings of the 2018 Conference on Empirical Methods in
  Natural Language Processing}, pages 4098--4109, 2018.

\end{thebibliography}

\end{document}